\pdfoutput=1
\documentclass[11pt]{article}

\usepackage[]{acl}

\usepackage{graphicx}
\usepackage{algpseudocode}
\usepackage{algorithm}

\usepackage{booktabs}
\usepackage{multirow}

\usepackage{times}
\usepackage{latexsym}

\usepackage[T1]{fontenc}
\usepackage[utf8]{inputenc}

\usepackage{microtype}

\title{Look Ma, Only 400 Samples! Revisiting the Effectiveness of Automatic N-Gram Rule Generation for Spelling Normalization in Filipino}

\author{Lorenzo Jaime Yu Flores \quad Dragomir Radev \\
Yale University \\
  \texttt{lj.flores@yale.edu}
}

\begin{document}
\maketitle
\begin{abstract}
With 84.75 million Filipinos online, the ability for models to process online text is crucial for developing Filipino NLP applications. To this end, spelling correction is a crucial preprocessing step for downstream processing. However, the lack of data prevents the use of language models for this task. In this paper, we propose an N-Gram + Damerau-Levenshtein distance model with automatic rule extraction. We train the model on 300 samples, and show that despite limited training data, it achieves good performance and outperforms other deep learning approaches in terms of accuracy and edit distance. Moreover, the model (1) requires little compute power, (2) trains in little time, thus allowing for retraining, and (3) is easily interpretable, allowing for direct troubleshooting, highlighting the success of traditional approaches over more complex deep learning models in settings where data is unavailable.
\end{abstract}

\section{Introduction}

Filipinos are among the most active social media users worldwide \citep{Baclig_2022}. In 2022, roughly 84.75M Filipinos were online \citep{statista_internet}, with 96.2\% on Facebook \citep{statista_fb}. Hence, developing language models that can process online text is crucial for Filipino NLP applications. 

Contractions and abbreviations are common in such online text \citep{Salvacion_2022}. For example, \textit{dito} (here) can be written as \textit{d2}, or \textit{nakakatawa} (funny) as \textit{nkktawa}, which are abbreviated based on their pronunciation. However, language models like Google Translate remain limited in their ability to detect and correct such words, as we find later in the paper. Hence, we aim to improve the spelling correction ability of such models.

In this paper, we demonstrate the effectiveness of a simple n-gram based algorithm for this task, inspired by prior work on automatic rule generation by \citet{Mangu_1997}. Specifically, we (1) create a training dataset of 300 examples, (2) automatically generate n-gram based spelling rules using the dataset, and (3) use the rules to propose and select candidates. We then demonstrate that this model outperforms seq-to-seq approaches.

Ultimately, the paper aims to highlight the use of traditional approaches in areas where SOTA language models are difficult to apply due to limitations in data availability. Such approaches have the added benefit of (1) requiring little compute power for training and inference, (2) training in very little time (allowing for frequent retraining), and (3) giving researchers full clarity over its inner workings, thereby improving the ease of troubleshooting.

\section{Related Work}

The problem of online text spelling correction is most closely related to \textit{spelling normalization} – the subtask of reverting shortcuts and abbreviations into their original form \citep{Nocon_2014}. In this paper, we will use \textit{correcting} to mean \textit{normalizing} a word. This is useful for low-resource languages like Filipino, wherein spelling is often not standardized across its users \citep{li_2020}.

Many approaches have been tried for word normalization in online Filipino text: (1) \textit{predetermined rules} using commonly seen patterns \citep{Guingab_2014, oco-borra-2011-grammar}, (2) \textit{dictionary-substitution models} for extracting patterns in misspelled words \citep{Nocon_2014}, or (3) \textit{trigrams} and \textit{Levenshtein or QWERTY distance} to select words which share similar trigrams or are close in terms of edit or keyboard distance \citep{Chan_2008, Go_2017}. 

Each method has its limitations which we seek to address. \textit{Predetermined rules} must be manually updated to learn emerging patterns, as is common in the constantly evolving vocabulary of online Filipino text \citep{Salvacion_2022, Lumabi_2020}. \textit{Dictionary-substitution models} are limited by the constraint of picking mapping each pattern to only a single substitution, whereas in reality, different patterns may need to be applied to different words bearing the same pattern \citep{Nocon_2014}. \textit{Trigrams and distance metrics} alone may be successful in the context of correcting typographical errors for which the model was developed \citep{Chan_2008}, but may not be as successful on intentionally abbreviated words. Our work uses a combination of these methods to develop a model that can be easily updated, considers multiple possible candidates, and works in the online text setting.

The task is further complicated by the lack of data, which hinders the use of large pretrained language models. Previous supervised modeling approaches require thousands of labeled examples \citep{etoori-etal-2018-automatic}, and even unsupervised approaches for similar problems required vocabulary lists containing the desired words for translation \citep{Conneau_2017, lample-etal-2018-phrase}. Since such datasets are not available, our paper revisits simpler models, and finds that they exhibit comparable performance to that of much larger SOTA models.

\section{Data}

We use a dataset consisting of Facebook comments made on weather advisories of a Philippine weather bureau in 2014. We identified 403 distinct abbreviated and contracted words, and had three Filipino undergraduate volunteers write their ``correct'' versions. To maximize the data, we removed hyphens and standardized spacing, then filtered out candidates where all annotators gave different answers. 

We obtained 398 examples (98.7\%) with 83.8\% inter-annotator agreement. We then created a 298-100 train-test split; we selected test examples that used spelling rules present in the training set to test the ability of our n-gram model to extract and apply such rules. To test generalizability, we also perform cross-validation. The data and code for our experiments are available at the following repository.   \footnote{\url{https://github.com/ljyflores/Filipino-Slang}}

\section{Model}

\paragraph{Automatic Rule Generation} We extract spelling rules from pairs $(w,c)$, where $w$ is a misspelled word, and $c$ is its corrected version. The rule generation algorithm slides a window of length $k$ over $w$ and $c$, and records $w[i:i+k] \rightarrow c[j:j+k]$ as a rule ($i,j$ are pointers); it returns a dictionary mapping each substring to a list of ``correct" substrings (See Appendix \ref{alg:rule_gen} for algorithm and example). 

We test substrings of length 1 to 4, and find that lengths 1 / 2 work best. This makes sense as many Filipino words are abbreviated by syllable, which typically have 1-2 letters. This is similar to Indonesian \citep{Batais2015} and Malay \citep{Ramli2015}, suggesting possible extensions.

We further filter candidates to words present in a Filipino vocabulary list developed by \citet{Gensaya_2018} (MIT License), except for when none of the candidates exist in the vocabulary list, in which case we use all the generated words as candidates.

\paragraph{Candidate Generation} We recursively generate candidates by replacing each substring with all possible rules in the rule dictionary. If the substring is not present, we keep the substring as is. An example can be found in Appendix \ref{appendix:example}.

We find that rules involving single letter substrings often occur at the end of a word. Hence, we test candidate generation algorithms which either allow single letter rules to be used anywhere when generating (V1), or only for the last letter of a word (V2). We also vary the \# of candidates kept at each generation step (ranked by likelihood, see Eq \ref{Eq:Likelihood}).

\paragraph{Ranking Candidates} We explore two ways of ranking candidates: (1) \textit{Damerau-Levenshtein Distance} we rank candidates based on their edit distance from the misspelled word using the pyxdameraulevenshtein\footnote{\url{https://github.com/lanl/pyxDamerauLevenshtein}} package with standard settings, and (2) \textit{Likelihood Score} we compute the likelihood of the output word $c$ given misspelled word $w$ as the product of probability the rules used to generate it, where the probability of a rule is the number of occurrences of $a \rightarrow b$ divided by the number of rules starting with $a$ (See Eqs \ref{Eq:Prob}, \ref{Eq:Likelihood}).

\begin{equation}
    \label{Eq:Prob}
    P(a \rightarrow b) = \frac{|\{a \rightarrow b\}|}{|\{a \rightarrow c\} \forall c|}
\end{equation}

\begin{equation}
    \label{Eq:Likelihood}
    \small P(w \rightarrow c) = \prod_{i=1}^{len(w)-k} P(w[i:i+k] \rightarrow c[i:i+k])
\end{equation}

\section{Evaluation}

\subsection{Comparison to Language Models} 

We benchmark the performance of our models against two seq-to-seq models on the same dataset: (1) \textbf{ByT5} \citep{xue-etal-2022-byt5}: a character-level T5 model \citep{Colin_2020} trained on cross-lingual tasks, shown to be robust to misspellings, and (2) \textbf{Roberta-Tagalog} \citep{cruz2021improving}: a BERT \citep{devlin-etal-2019-bert} model trained on large Filipino corpora for masked language modeling. We performed hyperparameter tuning using an 80-20 split of the training data.

For inference, we obtain the top five candidates for each misspelled word by selecting the highest scoring candidates using beam search.

\subsection{Augmentation Techniques} 

Since deep learning models perform poorly on small datasets, we use two techniques to improve performance to achieve more quality benchmarks. 

\begin{figure}
    \centering
    \includegraphics[width=1\columnwidth]{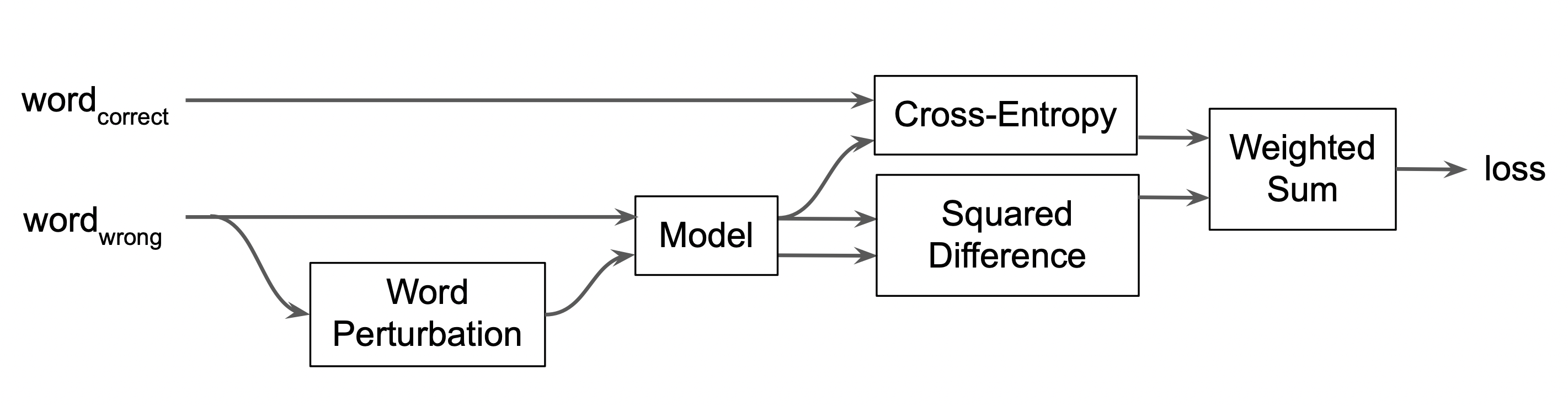}
    \caption{$\Pi$-Model Architecture
    \citep{Laine_2017}}
    \label{Fig:pi_model}
\end{figure}

First, we use $\Pi$-model \citep{Laine_2017} (Fig \ref{Fig:pi_model}), a semi-supervised technique which minimizes the mean-squared distance between the predicted corrections for two versions of a misspelled word, where the weight is a hyperparameter. % The final loss is a weighted sum of the seq2seq negative cross-entropy loss and the mean-square loss, 

Then, we use autoencoding augmentation (AE) \citep{bergmanis-etal-2017-training}, where we iteratively train a seq2seq model on the original spelling normalization task and an autoencoding task, where the model is trained to reproduce the same word.  % The loss is a weighted sum of the cross-entropy loss from both the correction and autoencoding task.

\subsection{Comparison to Google Translate} 

We also benchmark using Google Translate's model. We input each word and check if the model outputs a valid translation or suggests a correction (i.e. ``Did you mean X?"). A correct translation/correction means the model was able to correct (and thus translate) the misspelled word.

\subsection{Evaluation} 

We evaluate the models with two metrics: (1) \textbf{Accuracy @ k:} \% of observations where the target is present among the top-k candidates, and (2) \textbf{Damerau-Levenshtein Distance (DLD):} Best, average, and worst-case DLD of the top 5 candidates.

\section{Results}

\subsection{Results from Evaluation Metrics}

\begin{table*}
\centering
\begin{tabular}{llcccccc}
\toprule
\multirow{2}{0pt}{}Type & Model & 
    \multicolumn{3}{c}{Accuracy @ k (\%)} & 
    \multicolumn{3}{c}{DLD} \\
 & & $k=1$ & $k=3$ & $k=5$ & Min & Mean & Max \\
\midrule
N-Gram Based & N-Grams + DLD V1     & \textbf{0.77} & \textbf{0.82} & \textbf{0.85} & \textbf{0.46} & 2.91 & 4.73 \\
  & N-Grams + DLD V2     & 0.67 & 0.74 & 0.74 & 1.03 & 2.96 & 4.59 \\
 & N-Grams + Likelihood V1 & 0.17 & 0.38 & 0.58 & 1.22 & 3.50 & 5.29 \\
 & N-Grams + Likelihood V2 & 0.47 & 0.61 & 0.64 & 1.30 & 3.06 & 4.65 \\
ByT5 & Model Only                 & 0.31 & 0.42 & 0.49 & 0.98 & 2.71 & 4.38 \\
 & Model + $\Pi$-Model   & 0.37 & 0.58 & 0.66 & 0.57 & \textbf{2.06} & 3.41 \\
 & Model + AE   & 0.04 & 0.04 & 0.04 & 4.28 & 6.69 & 10.2 \\
Roberta-Tagalog & Model Only   & 0.00 & 0.00 & 0.00 & 5.79 & 15.3 & 56.7 \\
 & Model + $\Pi$-Model   & 0.00 & 0.00 & 0.00 & 5.69 & 16.5 & 69.2 \\
 & Model + AE   & 0.00 & 0.00 & 0.00 & 9.44 & 42.8 & 81.7 \\
Baselines & DLD                  & 0.45 & 0.67 & 0.72 & 0.59 & 2.28 & \textbf{3.32} \\
 & Google Translate     & 0.44 & -    & -   & -    & -    & -     \\
\bottomrule
\end{tabular}
\caption{\label{Table:Results}
Performance of Spelling Normalization Models on Test Set, see Appendix \ref{Table:Hyperparameters} for hyperparameter settings
}
\end{table*}

We train our models and show the results on the test set in Table \ref{Table:Results} (See Appendix \ref{Table:Hyperparameters} for hyperparameter details). To test generalizability, we perform 5-fold cross-validation (See Appendix \ref{Table:Results_CV}).

The N-Grams + DLD V1 algorithm performs best in terms of accuracy and best-case DLD. It achieves an improvement of 32\% from the next best model (DLD) for accuracy @ 1, which we consider most important, as real-world spellcheckers usually suggest one word. In addition, the ByT5 + $\Pi$-Model exhibits the best average DLD; hence it generates many candidates which resemble the target, though not achieving the correct output.

Also, N-Grams + Likelihood performs much worse than with DLD, despite using the same candidate generation procedure. This was because the dictionary also had irrelevant rules which muddled the estimates; these can be filtered out with heuristics, though at the expense of generalizability.

Moreover, the $\Pi$-model results in small improvements over the original ByT5 across all metrics; this illustrates the impact of semi-supervised approaches over supervised approaches in settings with limited data, albeit with limited success.

\subsection{N-Gram Algorithm Runtime}

Though N-Grams + DLD V1 achieves the best performance, it performs inference in 2.781s on average. N-Grams DLD V2 achieves significantly faster performance (0.0086s) with a marginal decrease in performance (See Appendix \ref{appendix:runtime}). It is worth noting that all N-Gram models train in under a second on a local CPU, whereas most language models required at least 20 minutes on a GPU.

% In contrast, the ByT5 and ByT5 + $\Pi$-Model required GPUs, though with faster inference time. % are there more details on amount of GPU memory/VRAM required for a single inference? / more info you can give on "requires GPUs"? 

\subsection{Analysis of Errors: N-Gram + DLD V1} 

We analyze the examples in which the N-Gram + DLD did not select the correct word as the top choice (i.e. error at $k=1$). The N-Gram + DLD model produced errors on 23 observations (out of 100); we separate these errors into those where the target was and was not in the candidate list.

\paragraph{Errors with Target in the Candidates} There were 9 (out of 23) errors wherein the target was not among the candidates. In such cases, the DL score selected candidates which closely resembled the input, but were wrong; the correct choices were ranked in the top 12.65\% of candidates on average (median of 8.57\%). Given the difficulty in distinguishing between words with similar spellings, context may be required (e.g. words surrounding the misspelled words, likelihood of word occurring).

\label{Section:Error_with_Ranking}

\paragraph{Errors with Target not in the Candidates} There were 14 (out of 23) errors with targets not in the candidate list; here, the rule dictionary lacked at least one rule that was necessary to correct each of the misspelled words. Upon adding these rules to the dictionary, the model correctly predicted all but five observations. In those five cases, the target was in the candidate list but not selected as the top result, suggesting the need for better ranking methods as discussed in the previous section.

As demonstrated by this section, a benefit of the N-Gram + DLD model is that it allows access to the collected rules, allowing researchers to understand the cause of such errors, and hence directly make tweaks (e.g. by adding rules, tweaking substring length $k$) to improve the model. In contrast, explainability remains a challenge for language models, thereby reducing their ease of troubleshooting.

\section{Conclusion}

In this study, we propose an N-Gram + DLD model for spelling normalization of Filipino online text, and compare it to deep learning benchmarks. The N-Gram + DLD V1 model achieves the best accuracy and best-case DLD, with a 32\% improvement in accuracy @ 1 over the next best model (DLD). This shows the potential of simpler techniques, especially when data is scarce.

Besides improved performance, the N-Gram + DLD model requires little compute power and memory for training and inference. This allows for frequent retraining of the model and addition of new spelling rules as new words emerge. The model also allows researchers to understand how predictions are made, and make appropriate tweaks to the spelling rules, candidate sorting method, or hyperparameters used (e.g. length of substrings).

This work has limitations which suggest areas for improvement. First, the current work uses a small dataset limited to the weather domain. Using more diverse datasets can improve the comprehensiveness of the rule dictionary. Also, more complete dictionaries containing Filipino words and their conjugations can help filter down valid candidates before running DLD. 

Second, the candidate ranking method can be improved, especially in cases where the target and selected words are similar, as discussed in the section \ref{Section:Error_with_Ranking}. For example, words can be ranked by how common they are, or by inferring the correct choice from the context. This has the added benefit of reducing the candidate pool, requiring fewer DLD calculations and hence reducing inference time. 

Finally, we only explore correcting misspelled words; combining it with misspelling detection can further boost the practical applications of this work.

Ultimately, the development of such models will pave the way for improvements in Filipino NLP, and enable the development of more applications that can serve the wider online Filipino community.

\section*{Acknowledgements}

We would like to thank Luis Angelo Chavez, Agnes Robang, and Mirella Arguelles for annotating the dataset, and Hailey Schoelkopf and Linyong Nan for their feedback on the paper.

\bibliography{anthology,custom}
\bibliographystyle{acl_natbib}

\appendix

\section{Cross Validation Results}

The cross validated results are shown in Table \ref{Table:Results_CV}. While the metrics dropped from that in Table \ref{Table:Results}, the models still exhibit the same order of performance in terms of accuracy.

\begin{table*}
\centering
\begin{tabular}{lrrrrrr}
\toprule
\multirow{2}{0pt}{} Model &
    \multicolumn{3}{c}{Accuracy @ k (\%)} & 
    \multicolumn{3}{c}{DLD} \\
  & $k=1$ & $k=3$ & $k=5$ & Min & Mean & Max \\
\midrule
RT & 0.0 ± 0.00 & 0.0 ± 0.00 & 0.0 ± 0.00 & 6.06 ± 0.55 & 12.0 ± 2.85 & 46.2 ± 20.0 \\
RT + $\Pi$ & 0.0 ± 0.00 & 0.0 ± 0.00 & 0.0 ± 0.00 & 6.08 ± 0.56 & 15.3 ± 2.77 & 61.7 ± 17.5 \\
RT + AE & 0.0 ± 0.00 & 0.00 ± 0.00 & 0.00 ± 0.00 & 7.38 ± 1.53 & 21.3 ± 6.92 & 54.9 ± 8.10 \\
BT & 0.32 ± 0.06 & 0.52 ± 0.05 & 0.59 ± 0.07 & 0.77 ± 0.15 & 2.31 ± 0.16 & 3.76 ± 0.26 \\
BT + $\Pi$ & 0.40 ± 0.06 & 0.57 ± 0.03 & 0.65 ± 0.03 & 0.53 ± 0.05 & 1.75 ± 0.07 & 2.83 ± 0.12 \\
BT + AE & 0.02 ± 0.03 & 0.02 ± 0.03 & 0.02 ± 0.03 & 4.05 ± 0.41 & 6.33 ± 0.38 & 9.45 ± 0.71 \\
NG + DLD & 0.53 ± 0.02 & 0.63 ± 0.04 & 0.65 ± 0.06 & 1.49 ± 0.11 & 2.93 ± 0.07 & 4.18 ± 0.11 \\
NG + Lik. & 0.35 ± 0.07 & 0.47 ± 0.08 & 0.49 ± 0.07 & 1.69 ± 0.26 & 2.95 ± 0.11 & 4.13 ± 0.16 \\
\bottomrule
\end{tabular}
\caption{\label{Table:Results_CV}
Five-fold cross validation results for models on joint train and test set, within one standard deviation Legend: RT (Roberta-Tagalog), BT (ByT5), NG (N-Grams)
}
\end{table*}

\section{Algorithms}

\begin{algorithm}[h]
\caption{Automatic Rule Generation} \label{alg:rule_gen}
\textbf{Input} $w$ (wrong word), $r$ (right word)\\
\textbf{Output} $d$ (rule dictionary)
\begin{algorithmic}[1]
\State $k, d\gets \{\}, ptr_{w}=0, ptr_{r}=0$
\While{$ptr_{w}<len(w) \ \& \ ptr_{r}<len(r)$} % \Comment{Continue if we have not reached end of either string}
    \State $substr_w \gets w[ptr_w:ptr_w+k]$
    \State $substr_r \gets r[ptr_r:ptr_r+k]$
    \If{$substr_w = substr_r$} 
        \State $ptr_w \gets ptr_w + k$ 
        \State $ptr_r \gets ptr_r + k$
    \Else
        \State $ptr_w \gets ptr_w + 1$ 
        \State $ptr_r \gets ptr_r + k$ 
    \EndIf 
    \State Append $substr_r$ to key $substr_w$ in $d$
\EndWhile
\State \textbf{Return} $d$
\end{algorithmic}
\end{algorithm}

\begin{figure}[h]
    \centering
    \includegraphics[width=1\columnwidth]{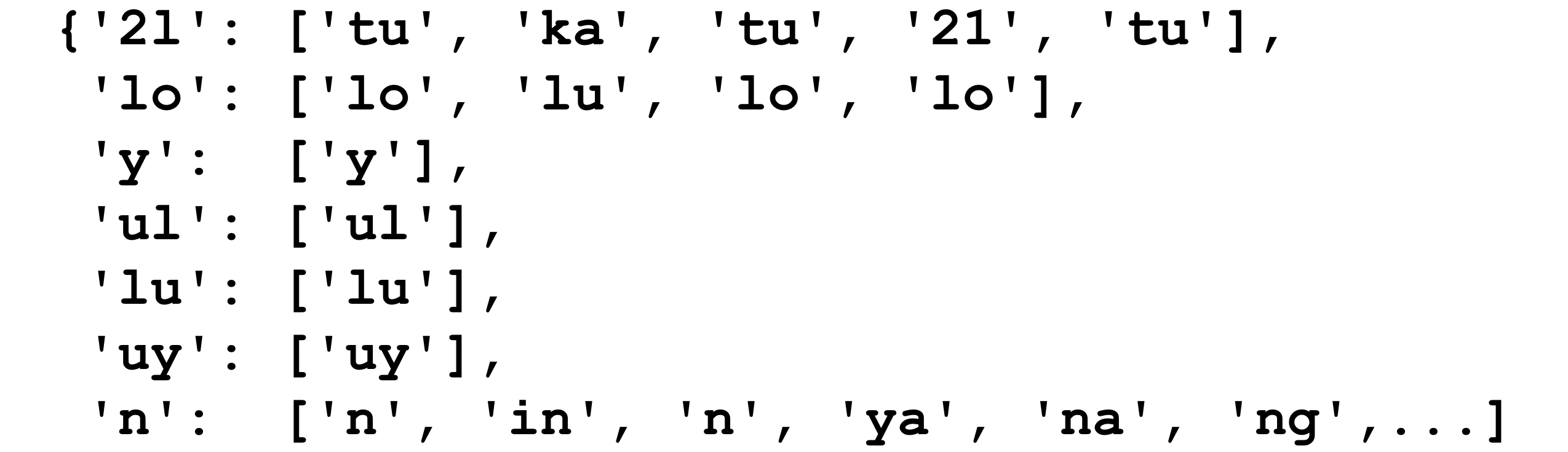}
    \caption{Example of a generated rule dictionary}
    \label{Fig:example_dictionary}
\end{figure}

\section{Runtime Performance}
\label{appendix:runtime}

\begin{figure}[h]
    \centering
    \includegraphics[width=1\columnwidth]{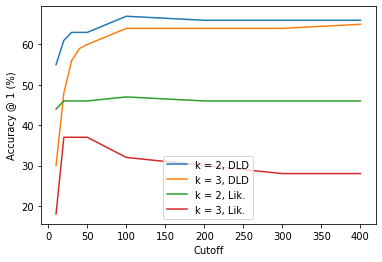}
    \caption{N-Gram Cutoff vs. Test Set Accuracy @ 1 (\%), $k$ is the maximum length of substring considered}
    \label{Fig:cutoff_vs_acc}
\end{figure}

\begin{figure}[t!]
    \centering
    \includegraphics[width=1\columnwidth]{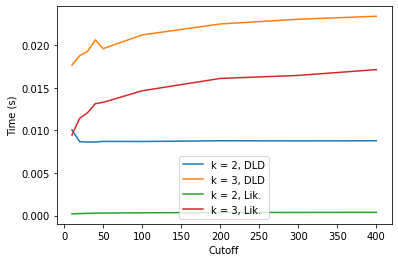}
    \caption{N-Gram Cutoff vs. Inference Time (s), $k$ is the maximum length of substring considered}
    \label{Fig:cutoff_vs_time}
\end{figure}

We plot accuracy @ 1 and runtime in Figures \ref{Fig:cutoff_vs_acc} and \ref{Fig:cutoff_vs_time} respectively, and find that using a cutoff of 100 and 30 for N-Grams + DLD and N-Grams + Likelihood respectively achieve the best tradeoff between runtime and performance.

\iffalse

\begin{algorithm}
\caption{Candidate Generate}\label{alg:cand_gen}
\textbf{Input} $w$ (wrong word), $d$ (rule dict), $k$\\
\textbf{Output} \text{List of candidates}

\begin{algorithmic}[1]
\Procedure{Generate}{$w,d,k$}
\State $k, d\gets \{\}, ptr_{w}=0, ptr_{r}=0$
\If{$len(w)=0$}
    \State \textbf{Return} $[``"]$
\EndIf
\State $Results \gets []$
\For{$i=1,2,\ldots,k$}
    \If{$len(w)<k$}
        \State $continue$
    \EndIf
    \State $substr \gets w[:k]$
    \If{$substr \in d$}
        \State $replacements \gets d[substr]$
        \For{$r \in replacements$}
            \If{$r=substr$}
                \State $next \gets \text{Generate}(w[k:], d, k)$
            \Else{$next \gets \text{Generate}(w[1:], d, k)$}
            \EndIf
            \State $combos \gets [r+i \ \forall \ i \in next]$
            \State Add $combos$ to $Results$
        \EndFor
    \Else{}
        \State $next = \text{Generate}(w[1:], d, k)$
        \State $temp = [w[0]+i \ \forall \ i \in next]$
        \State Add $temp$ to $Result$
    \EndIf
    
\EndFor

\State \textbf{Return} $Results$
\end{algorithmic}
\end{algorithm}
\fi

\section{Example}
\label{appendix:example}
\begin{figure}[!t]
    \centering
    \includegraphics[width=1\columnwidth]{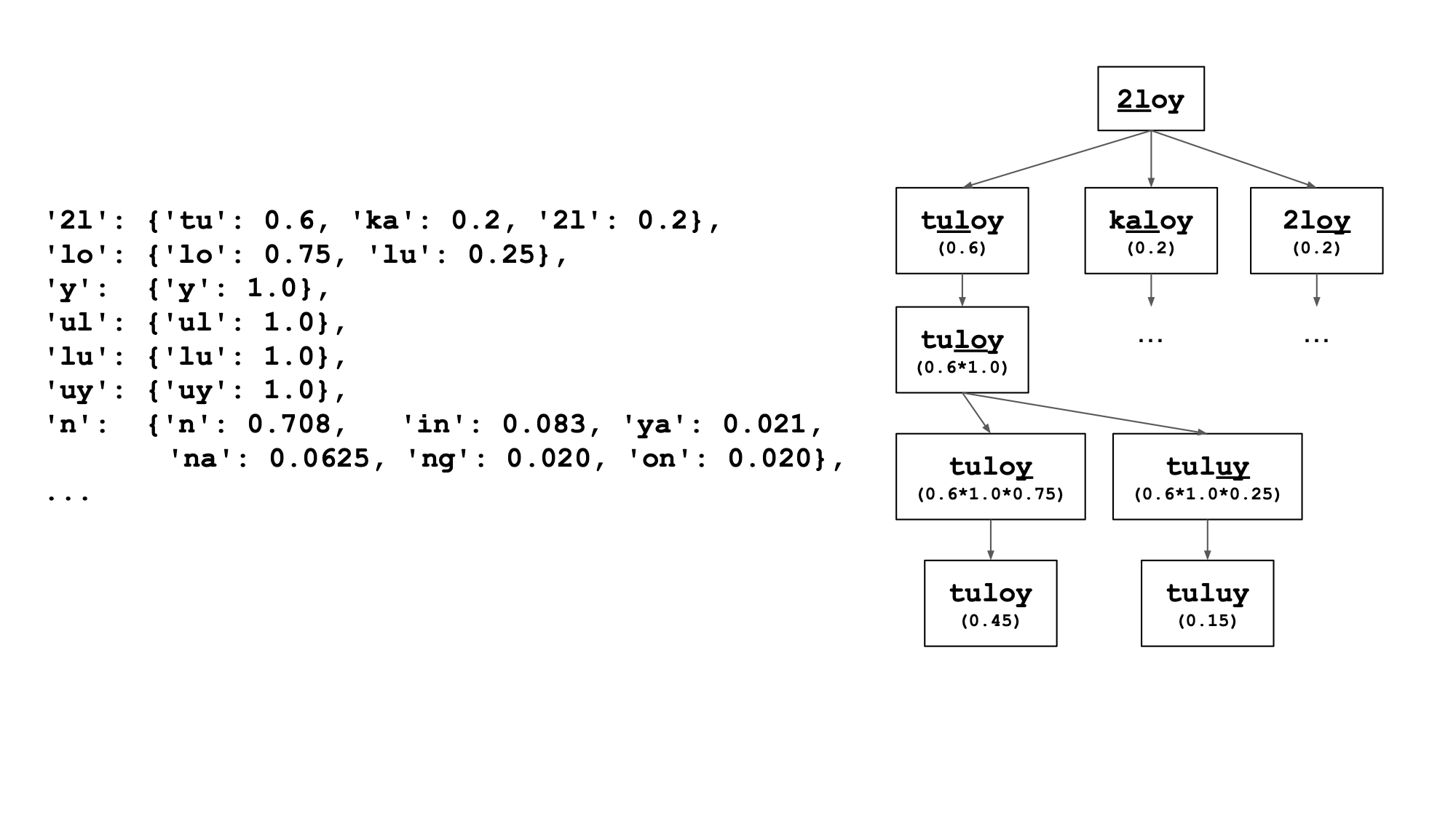}
    \caption{Example Inference for \textit{``2loy''}}
    \label{Fig:example}
\end{figure}

Figure \ref{Fig:example} shows a rule dictionary and how the rules are used to normalize \textit{``2loy''} to \textit{``tuloy''}.

\section{Computational Details}

We use one RTX 3090 (24GiB) GPU to perform training for the language models, and we used a total of six GPU hours across finetuning and hyperparameter selection. We note that ByT5 consists of 300 million parameters.

\section{Hyperparameter Settings}
We train all models with the Adam optimizer, with a starting learning rate of $5\mathrm{e}{-5}$ and stability of $1\mathrm{e}{-8}$. Hyperparameters were finetuned using Ray Tune, and models were selected based on the lowest validation loss, as shown in Table \ref{Table:Hyperparameters}.

\begin{table}[h!]
\centering
\begin{tabular}{lrrr}
\toprule
Model & Batch & Epochs & MSE Weight \\
\midrule
RT & 8 & 10 & - \\
RT + $\Pi$ & 8 & 30 & 0.2 \\
RT + AE & 4 & 70 & - \\
BT & 1 & 50 & - \\
BT + $\Pi$ & 1 & 70 & 0.2 \\
BT + AE & 4 & 70 & - \\
\bottomrule
\end{tabular}
\caption{\label{Table:Hyperparameters}
Hyperparameter settings for best models, finetuned using the Ray Tune Python library; We tried 10, 30, 50, 70 epochs, and batch sizes of 1, 2, 4, 8, and 16; Legend: RT (Roberta-Tagalog), BT (ByT5), NG (N-Grams)
}
\end{table}
\end{document}